\documentclass[a4paper, times,twocolumn,
authoryear]{article}
\usepackage[a4paper, margin=0.6in]{geometry}

\usepackage{framed,multirow}

\usepackage{amssymb}
\usepackage{latexsym}
\usepackage{natbib}

\usepackage{url}
\usepackage{xcolor}
\usepackage{tikz}

\definecolor{newcolor}{rgb}{.8,.349,.1}
\usepackage{subcaption}

\begin{document}

\author{
Anatoly Belikov$^1$ 
\and
Alexey Potapov$^1$
}

\date{%
    $^1$SingularityNet\\%
    \today
}

\title{GoodPoint: unsupervised learning of keypoint detection and description\footnote{the paper is under consideration at Pattern
Recognition  Letters}
}






\maketitle

\begin{abstract}
\emph{
This paper introduces a new algorithm for unsupervised learning of keypoint detectors and descriptors, which demonstrates fast convergence  and good performance across different datasets. The training procedure uses homographic transformation of images. The proposed model learns to detect points and generate descriptors on pairs of transformed images, which are easy for it to distinguish and repeatedly detect. The trained model follows SuperPoint architecture for ease of comparison, and demonstrates similar performance on natural images from HPatches dataset, and better performance on retina images from Fundus Image Registration Dataset, which contain low number of corner-like features. For HPatches and other datasets, coverage was also computed to provide better estimation of model quality.
}

\end{abstract}


\section{Introduction}
\label{sec1}

Local image features (keypoint detection and descriptor extraction) form the base of many computer vision applications, most notably simultaneous localization and mapping (SLAM) and augmented reality. Traditionally, handcrafted local features were used such as Harris corner detector  \citep{harris1988combined}, SURF \citep{funayama2012robust} and many others, but machine learning methods have shown their usefulness for the task quite early. For example, FAST \citep{rosten2006machine} introduced in 2006 uses decision trees for corner detection. With the improvement of hardware and deep learning theory, it became also possible to learn descriptor extraction and matching, as for example in SuperGlue \citep{sarlin2019superglue}. However, most typically, supervised learning is used that limits the applicability of the methods to novel domains.

One of the definitions of machine learning is the ability of a program to improve performance with more data \citep{mitchell1997machine}. Deployed feature extraction and image matching methods are to be applied to unlabelled data, and the improvement of their performance naturally supposes unsupervised learning. Although supervised learning has been frequently demonstrating better performance, unsupervised training of convolutional neural networks for feature generation \citep{detone2018superpoint, truong2019glampoints} provides state-of-the-art results. Can this be achieved for the whole keypoint detection and description extraction pipeline?

The supervised method SuperPoint \citep{detone2018superpoint} features a very simple loss function for descriptors, which minimises the difference of descriptors of regions that correspond each other geometrically, and maximizes the difference otherwise. Since the heatmap for keypoints is a kind of descriptor too, it hints at the possibility of building good keypoint detectors in unsupervised manner with a simple loss function.
A popular type of keypoints among handcrafted or supervised detectors are corners. Among aforementioned methods, FAST, Harris and SuperPoint detect corners (and also line ends), SURF uses blob detection. 

From a practical point of view, an ideal keypoint detector is the one that optimizes performance of a downstream task (image matching) or even target application (e.g. SLAM), but this measure might be difficult to compute and/or optimize. Instead, we assume that good keypoints should possess the following properties:
\begin{enumerate}
\item
they should be distributed more or less evenly throughout the image;
\item
have good repeatability between different viewpoints;
\item
be recognizable and distinguishable with descriptors;
\item
should not lie too densely.
\end{enumerate}

In this paper, a new unsupervised algorithm for simultaneous training of the keypoint detector and the descriptor generator is proposed. A single two-headed neural network built up on SuperPoint architecture is used for both tasks. 
The proposed model can be trained directly on a target domain without the need for performing costly domain adaptation, and it is applicable in situations in which domain adaptation wouldn't work because of large difference between target and source domains. The proposed model achieves competitive performance with SuperPoint when trained on the same dataset, without supervised pre-training, and demonstrates better performance on images with low number of corner-like features. The resulting model is referred hereinafter as GoodPoint.

\section{Related work}

SuperPoint\citep{detone2018superpoint} introduced a fast convolutional neural network for keypoint detection and descriptor extraction. Training is split into two stages: 
\begin{enumerate}
\item supervised training of a detector on synthetic dataset;
\item training of a detector on self-labelled natural images together with unsupervised training of a descriptor. 
\end{enumerate}

Our work follows SuperPoint architecture, but simplifies training procedure, removing supervised pre-training and self-labelling from the pipeline. Instead, both heads of the network are trained on natural images from the beginning.

Authors of GLAMpoint\citep{truong2019glampoints} train a keypoint detector on pairs of images related by a homographic transformation. The method uses non-maximum suppression on heatmaps on both images to extract candidate keypoints and then uses matching with SURF descriptors \citep{funayama2012robust} to mine positive/negative examples. 

Another related research direction is object keypoint detection. Authors of \citep{jakab2018unsupervised} propose unsupervised keypoint-detector learning with conditional image generation. Given a pair of images $(x, x')$ with the same objects, but with a different viewpoint and/or object pose, a training procedure minimises weighted difference between features extracted from image $x'$ and the reconstruction $\hat x' = \Psi(x, K(x'))$ of $x'$. The reconstruction produced by a neural network $\Psi$, given image $x$ and keypoints from $x'$. The loss function defined as  
 $L = \sum_{l} \alpha_l ||\Gamma_l (x') - \Gamma_l( \hat x' )||^2_2$. Here $\Gamma_l$ denotes output of a layer $l$ of a pretrained neural network $\Gamma$. 

Here, $K$ is a keypoint detector neural network that learns to output $k$ heatmaps $K(x')=\mathbb{R}^{H \times W \times K}$, where $H$, $W$ are the height and the width of images $x$ and $x'$. Each heatmap corresponds to the location of one keypoint and is normalized with softmax function to be a probability distribution. 

Authors of \citep{kulkarni2019unsupervised} reuse the formulation from work\citep{jakab2018unsupervised} restricting to static backgrounds. The main difference from \citep{jakab2018unsupervised} is the introduction of feature transport: features extracted from both image used to generate $x'$: \newline
$\Phi(x,x') = (1-H(K(x)))·(1-H(K(x'))·\Phi(x) + H(K(x'))·\Phi(x')$

Here, $K$ is a keypoint detector that outputs k heatmaps. $H$ is a heatmap image containing isotropic Gaussians around each of the keypoints that are specified by $K(x)$ or $K(x')$. $\Phi$ is a feature extraction network. $\Phi$ takes background features (that is, features from locations where there are no keypoints) from both images plus features from $x'$ near target keypoints $K(x')$. Loss is squared reconstruction error:
$||x' - RefineNet(\Phi(x,x'))||^2_2$, where RefineNet is a convolutional generative model.

    LF-Net\citep{ono2018lf} advances the results of SuperPoint to the new state-of-the-art on many datasets, though it requires ground truth depth and camera pose information. LF-Net projects score map from source image $I_i$ to target image $I_j$, applies non-maximum suppression to sample keypoints and then generates new target score map with Gaussian kernel. The average difference of score maps is minimised:
$L(S_i,S_j)=|S_i - g(w(S_j))|_2$, where $w$ is a projection function, and $g$ is Gaussian kernel application.
Descriptors are extracted from keypoint locations, difference between correspondent, non-occluded descriptors is minimised.
Also, there is an additional loss for keypoint scales and orientations.

\section{Architecture overview}
\begin{figure*}[!t]
\centering
\includegraphics[scale=.35]{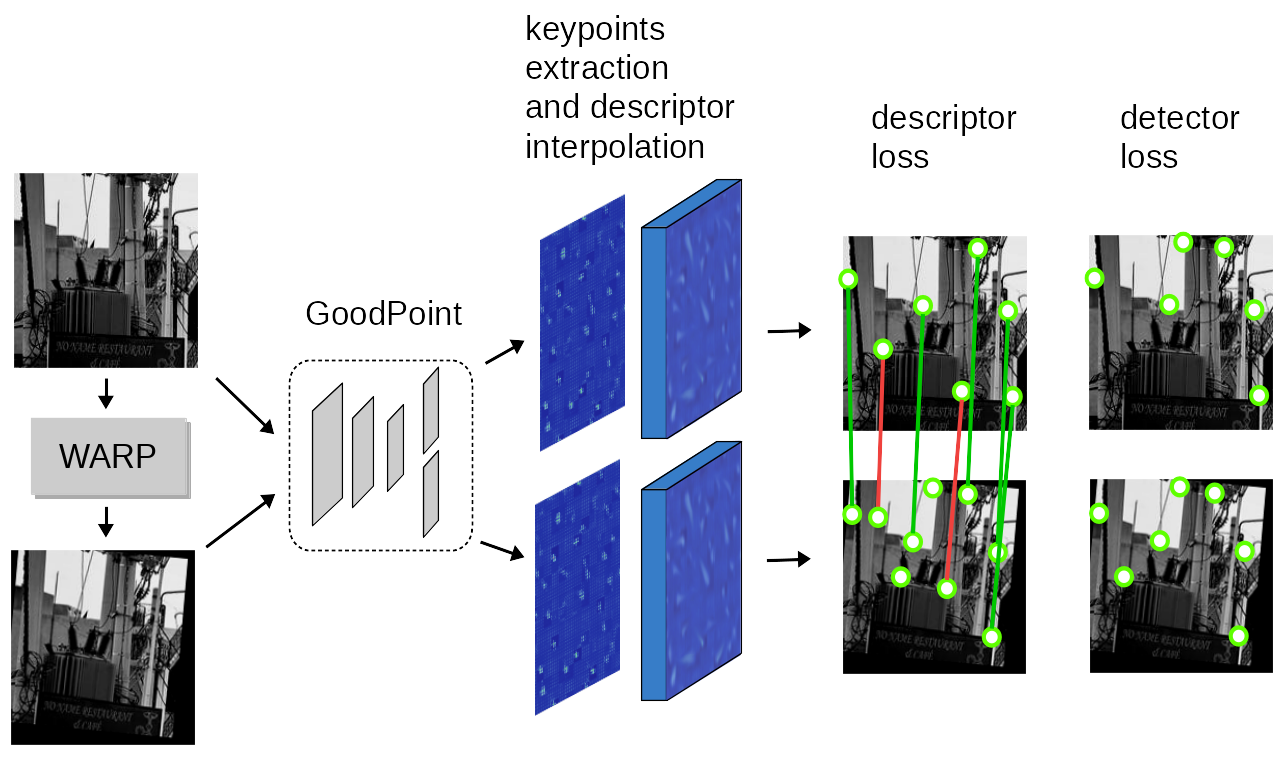} 
\caption{Unsupervised training overview.
First, keypoints and descriptors are extracted from original and warped images with a two-headed neural network. Descriptors are interpolated in location of keypoints from semi-dense output. Keypoints are matched with descriptors and correctly matched points are used as positive examples for detector training. All interpolated descriptors are used to calculate descriptor loss.
}
\label{fig:unsupervised_overview}
\end{figure*}

The proposed GoodPoint architecture is based on SuperPoint architecture and consists of a common VGG backbone followed by two heads: descriptor and detector. The VGG backbone and descriptor heads are left unchanged, except for the activation function. The training procedure, detector head and loss function are different. Activation function used for all layers is leaky ReLU\citep{maas2013rectifier}. So, the total number of trainable parameters is the same.
The detector is implemented similar to SuperPoint but without dustbin channel. So the detector head outputs tensor $P \in \mathbb{R} ^ {H/8 \times W/8 \times 64}$
 instead of $P \in \mathbb{R} ^{H/8 \times W/8 \times 65}$. This doesn't affect performance and simplifies implementation since all channels are now being treated equally. Softmax is applied along last axis to ensure that points lie not too densely. Also softmax makes it possible to learn only from positive examples. After the softmax, normalized tensors are reshaped from $\mathbb{R} ^ {H/8 \times W/8 \times 64}$ to $\mathbb{R}^ {H \times W \times 1}$ to form a confidence map.
The descriptor head outputs semi-dense tensor $D \in \mathbb{R} ^ {H/8 \times W/8 \times 256}$ which is interpolated in keypoint locations.

\section{Training}

Training is based on homographic warping of images and noise augmentation. The loss is computed on a pair of images: original $I$ and warped with random homography image $I_h$ .  Single homography $H$ is used for all images in a mini-batch. Both of the images may be warped, in which case they still are related by single homography $H$, so equations don't change. The final loss $L$ is a weighted sum of two losses, descriptor loss $L_d$  and detector loss 
\begin{equation}
L(P, Ph,D, Dh, H) =\lambda_1 L_d + \lambda_2 L_p
\end{equation}
where $\lambda_1$,  $\lambda_1$ are weights. $P$, $P_h$ are heatmaps for images $I$ and $I_h$.
$D$, $D_h$ are descriptors for images $I$ and $I_h$.

After homographic warp, random noise filters are applied independently to images $I$ , and $I_h$. Details of noise augmentation are provided in section \ref{sec-noise}.

\begin{figure}[!t]
\centering
\includegraphics[scale=.35]{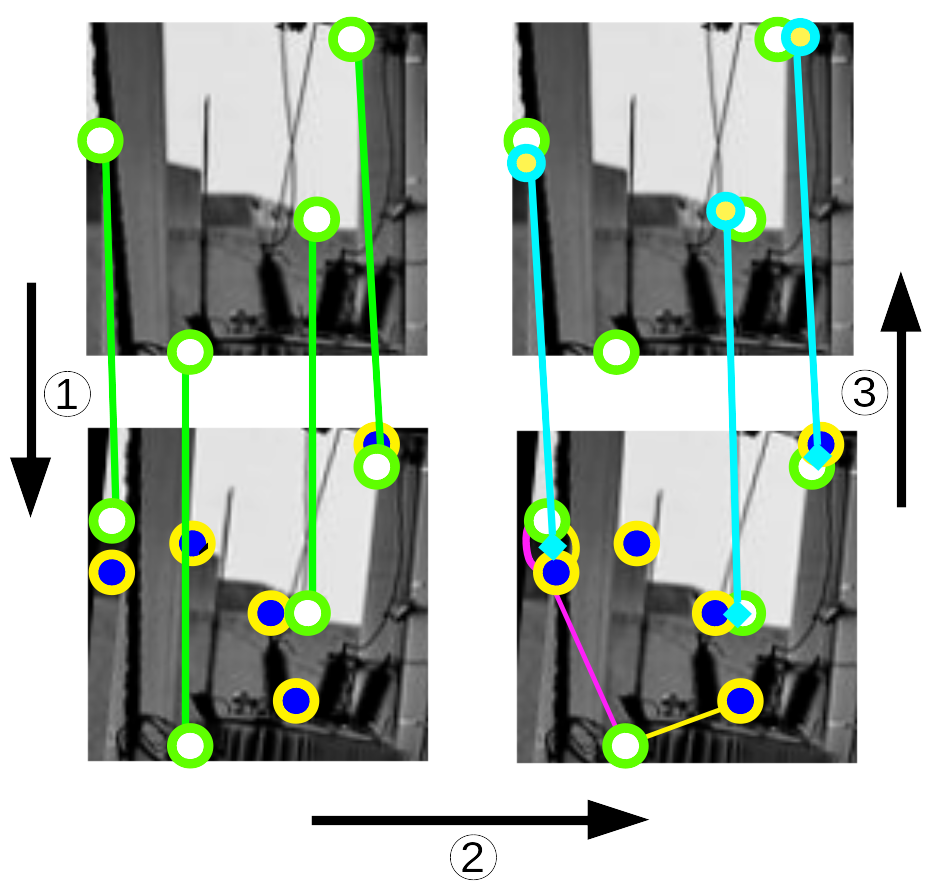} 
\caption[LoF entry]{keypoint target estimation

\tikz\draw[green,fill=white, very thick] (0,0) circle (0.7ex); - $K_{proj}$, \tikz\draw[cyan,fill=yellow, very thick] (0,0) circle (0.7ex);  - $K'_{h}$ ,$K'$

\tikz\draw[yellow,fill=blue, very thick] (0,0) circle (0.7ex);  - $K_{h}$

\protect \thicklines \textcolor{yellow}{ \line(2, 1){10} } - geometric match \hspace{1cm} \textcolor{green}{ \line(2, 1){10} } - projection from $I$ to $I_h$

\protect \thicklines \textcolor{magenta}{ \line(2, 1){10} } - descriptor match \hspace{1cm} \textcolor{cyan}{ \line(2, 1){10} } - projection from $I_h$ to $I$

}
\label{fig:target-est}
\end{figure}

Training of keypoint detection is inspired by expectation-maximisation technique. The network learns to output keypoints that are easy for it to reproduce.  It is trained with target keypoints computed with the following procedure (see figure \ref{fig:target-est}):
\begin{enumerate}
  \item Points $K$ found on image $I$ are projected to $I_h$ to form $K_{proj}$.
  \item Projected points $K_{proj}$ are matched with $K_h$ by 2D coordinates and by descriptors with the nearest neighbour matcher to form two sets of matches $K_{proj} \to K_h$.
  \item Pairs of points that match by coordinates and by descriptors (i.e. pairs are present in both sets of matches) as the nearest neighbours are used to compute targets. Targets $K'_h$ are projected back to image $I$.
\end{enumerate}

A more detailed description is provided in the next section.

\subsection{Keypoints loss}
The loss function for keypoint detector is a sum of negative log-likelihoods of estimated target keypoint positions for both images plus heatmaps difference:
\begin{equation}
L_p(P, Ph, K, Kh, H) = L_{keypoints} + L_{heatmaps}  \label{eq_loss_points}
\end{equation}

\begin{equation}
L_{keypoints}= - {1 \over  2} (log P[K'] + log P_h[K'_h])                    
\label{eq_loss_keypoints}
 \end{equation}   
\begin{equation}
L_{heatmaps} = \lambda_h {1 \over N_{mask}} 
\sum_{(i, j) \in mask } ^ {height, width}
(blr(\widehat{PH}) - blr(P_h I))^2(i,j) 
\end{equation}

$P[*]$ denotes selection of points * from 2D heatmap $P$.
$\lambda_h$ is a weight for heatmap difference.
$blr(\widehat{PH})$ denotes homographic projection of heatmap $P$, same way it is done for image $I$. Note that bilinear interpolation of $P_h$ is necessary, otherwise loss will be high even if heatmaps are similar due to $blr(\widehat{PH})$ being much smoother than $P$ or $P_h$. 

The sum iterates over points covered by mask for image $I_h$. Mask for image  $I_h$ is 2D tensor of the same shape the image, such that for all points $p=(x,y)$ in the mask: $mask[p] = 1$ if projection $\widehat{pH_{inv}} \in I_h$ and 0 otherwise.
$N_{mask}$ is the number of nonzero elements of the mask.
Given two heatmaps $P$ and $P_h$, tensors of estimated good keypoints positions $K'$ and $K'_h$ are computed with the following steps:
\begin{enumerate}
\item
Keypoint arrays $K$ and $K_h$ are extracted from $P$ and $P_h$  with maxpooling of different sizes:
\begin{equation}
K = maxpool_{32x32}(P)
\end{equation}
\begin{equation}
K_h = maxpool_{16x16}(P_h)
\end{equation}
 Selecting one keypoint for each region of size $32 \times 32$ or 
$16 \times 16$ follows from the assumption that keypoints should be distributed more or less evenly throughout an image, but not too densely.
Function maxpool performs maxpooling and returns coordinates of keypoints $(x_i, y_i)$  as an array. That is, $K$ and $K_h$ have shapes $ m \times 2 $  and $ n \times 2$ .

\item
Let the projection of points K to image plane $K_h$ be $K_{proj} = \widehat {KH}$, the keypoints projected beyond the boundaries of the image are discarded.

$D_{proj}$ , $D_h$ are descriptors of points $K_{proj}$ and $K_h$, i.e. $D_{proj}$ are descriptors extracted from image $I$ of keypoints that stay in bounds when projected on image $I_h$.

\item
The next step is to match points in $I_h$ with descriptors and with coordinates:
\begin{equation}
dist_{geom}, idx_{geom} = match_{geom}(K_{proj}, K_h)
\end{equation}
\begin{equation}
idx_{desc} = match_{desc}(D_{proj}, D_h)
\end{equation}
Here, the function $match_{geom}$ performs nearest neighbour matching of points $K_{proj}$ to $K_h$, with Euclidean distance between coordinates as a measure. 

$match_{geom}$ returns two vectors with length equal to length of $K_{proj}$. The first is the distance from a point in $K_{proj}$ to nearest point in $K_h$. The second gives an index of a nearest point in $K_h$. 

$idx_{desc}$ also gives an index of a nearest point in $K_h$, but with distance computed in the space of descriptors. So, $idx_{geom}$ and $idx_{desc}$ are of the same length.
\item
Positive examples for keypoints in image $I_h$ computed as mean coordinates of correctly matching points:

\begin{equation}
K'_h = coords_{mean}(K_{proj}(i), K_h[idx_{geom}(i)])
\end{equation} 
for $i$, such that  $idx_{geom}(i) == idx_{desc}(i)$ and $dist_{geom}(i) < \theta_{dist}$ i.e. indices should match, and geometric distance should be less than threshold.
Here $coords_{mean}(k_1, k_2) = 0.5  (k_1 + k_2)$. 
$\theta_{dist}$ is threshold in pixels for case that  distant points are matched correctly. 
$K'_h$ is then projected to image $I$ with inverse homography $H_{inv}$.
\begin{equation}
K' = K'_h H_{inv}
\end{equation}

\end{enumerate}
Thus we have targets $K'$ and $K'_h$  for both images that are needed to compute equation \ref{eq_loss_keypoints}.

\subsection{Descriptor loss}

Loss for descriptors consists of three components:
\begin{equation}
L_{desc}(D, D_h, K, K_h, H) = L_{gt} + L_{wrong} + L_{random} 
\end{equation}

Let element $i$ of vector $g_i = D_{proj}(i)D_h^T(idx_{geom}[i])$, i.e. scalar product of descriptors of keypoints matched by their coordinates, not by descriptors. Descriptors are normalized, so scalar product equals to cosine similarity. $L_{gt}$ maximises similarity of descriptors for each pair of points.
$L_{gt} = {1 \over N_{gt}} \sum_j (1 - g_j)$

$L_{wrong}$ minimises similarity of incorrectly matched pairs of descriptors, of points that are reasonably distant from each other.

\begin{equation}
L_{wrong}= {1 \over N_{wrong}} \sum_j g_j
\end{equation}
For such $j$, that $ idx_{geom}(j) \neq idx_{desc}(j) \land dist_{geom}(j) > 7$.
\newline
$L_{random}$ minimises difference of randomly sampled descriptors.

\begin{equation}
L_{random}= {1 \over N_{random}N_{points}} \sum_{i=0}^{N_{random}} \sum_{j=0}^{N_{points}} D_{proj}(i)sh(D_h(i))^T 
\end{equation}
$sh(D)$ is randomized shuffle of rows of descriptor matrix, such that no pair of $D_{proj}(i), D(i)$ would belong to nearest neighbours as defined by $idx_{geom}$.
\section{implementation details}
The model was implemented with pytorch framework. Optimization algorithm used during training is AdamW\citep{loshchilov2017decoupled} with initial learning rate of 0.0005, all other parameters are set to default values, particularly weight decay has default of 0.01. The proposed model was trained on the training set images from MS COCO dataset\citep{lin2014microsoft}. Each minimatch was composed from random crops of size 256x256 px. Weight for heatmap differenceh was set to 2000. The network was trained with constant learning rate for the first 8 epochs, after 8th epoch exponential decay of learning rate was used for 10 more epochs. 

\subsection{Noise augmentation}
\label{sec-noise}
Noise filters are applied in predefined order, sequentially to each image.
Each filter is skipped with probability 0.5.
Filters used during training:
\begin{enumerate}
\item
additive Gaussian
\item
random brightness
\item
additive shade
\item
salt \& pepper
\item
motion blur
\item
random contrast scale
\end{enumerate}
After each filter application, the image is checked for validity. Image is considered ruined if its variance is less than 10\%  of original in which case filter is skipped.

\subsection{Homographic augmentation}
\begin{figure}[!t]
\centering
\includegraphics[scale=.43]{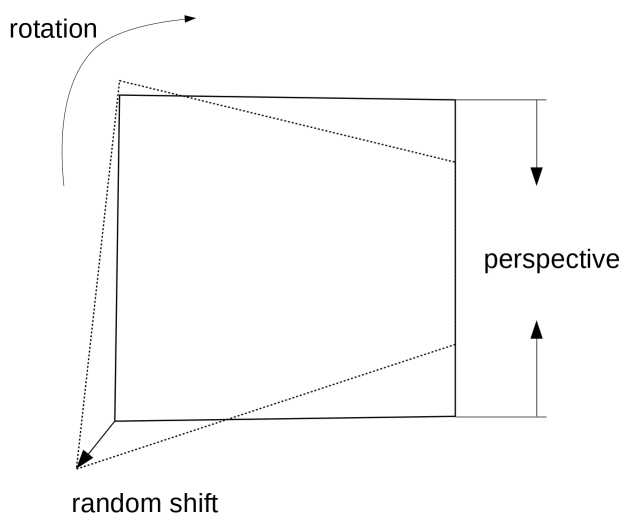} 
\caption{Random homography.
Homography is estimated from random perturbations of rectangle points.}
\label{fig:homography}
\end{figure}
Random homography matrices are generated as a product of simple transformations.
Random shift of points in range $\pm$14 px.
Perspective shift of side and/or top or bottom points in range 	$\pm$85 px.
Random homography augmentation was applied to both $I$ and $I_h$ with random rotation sampled from range $\pm$0.08 rad.
\subsection{Assessing performance}

\begin{table}[]
\caption{Test results on AirSim dataset}
\begin{tabular}{|l|l|l|}
\hline
                                                                           dataset
                                                                             & fantasy village                                                                                                                 & village                                                                                                                              \\ 
                                                                             \hline
SuperPoint                                                                   & 
\begin{tabular}[c]{@{}l@{}}
precision: 0.86 \\
repeatability: 0.57 \\
Coverage: 0.57\\
harmonic mean: 0.64
\end{tabular} & 
\begin{tabular}[c]{@{}l@{}}
precision: 0.72\\
repeatability: 0.45\\
Coverage: 0.65\\  
harmonic mean: 0.58
\end{tabular} 

\\ \hline
GoodPoint                                                                    & \begin{tabular}[c]{@{}l@{}}
precision: 0.85\\
Repeatability: 0.55\\ 
Coverage: 0.65\\ 
harmonic mean: 0.66
\end{tabular}  & 
\begin{tabular}[c]{@{}l@{}}precision: 0.74\\
repeatability: 0.42\\
Coverage: 0.70\\ 
harmonic mean: 0.58
\end{tabular}
\\ \hline
\begin{tabular}[c]{@{}l@{}}
SuperPoint\\ 
$5^{\circ}$ rotation
\end{tabular} 
& \begin{tabular}[c]{@{}l@{}}
Precision :0.85\\
Repeatability: 0.54\\
Coverage: 0.56\\
harmonic mean: 0.63
\end{tabular}  
& \begin{tabular}[c]{@{}l@{}}
Mean recall: 0.70\\
Repeatability: 0.42\\
Coverage: 0.62\\
harmonic mean: 0.55
\end{tabular}      
\\ \hline
\begin{tabular}[c]{@{}l@{}}GoodPoint\\ 
$5^{\circ}$ rotation\end{tabular}  & 
\begin{tabular}[c]{@{}l@{}}Precision: 0.85\\
Repeatability: 0.54\\
Coverage: 0.63\\ 
harmonic mean: 0.65
\end{tabular}    
& \begin{tabular}[c]{@{}l@{}}
Mean recall: 0.70\\
Repeatability: 0.39\\
Coverage: 0.67\\
harmonic mean: 0.55
\end{tabular}      \\ \hline
\end{tabular}

\label{table:airsim}
\end{table}
In a two-headed neural network, there is a trade-off between the performance of detector and descriptor networks. Computing a single metric that combines points repeatability and the precision of matching with descriptors is one way to break ties among multiple model variants. Authors of the paper [9] propose the following F1-like metric:
F1 = $2\times (\textrm{precision}(D, K)\times \textrm{repeatability}(K) / (\textrm{precision}(D, K) + \textrm{repeatability}(K))$ i.e. harmonic mean of precision of matching and keypoints repeatability, which was used for tuning hyper-parameters during training. Here, $D$ are descriptors, $K$ are keypoints. So, for all experiments we compute harmonic mean of all evaluation metrics, which gives a single number for comparison.

For all datasets, we also calculate coverage additionally to replication ratio and accuracy. The methodology was proposed in Irschara et al. \citep{irschara2009structure}.
Coverage is a ratio of covered pixels to all pixels in an image, with a pixel considered as covered when it lies within a certain distances from correctly matched keypoint.
\section{Experiments}
\begin{figure}[!t]
  \centering
  \begin{tabular}{@{}l l@{}}
    \includegraphics[scale=.295]{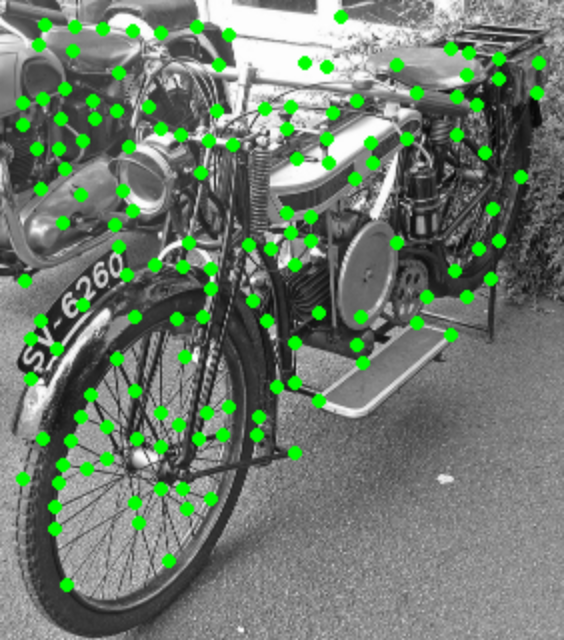} 
    & 
    \includegraphics[scale=.295]{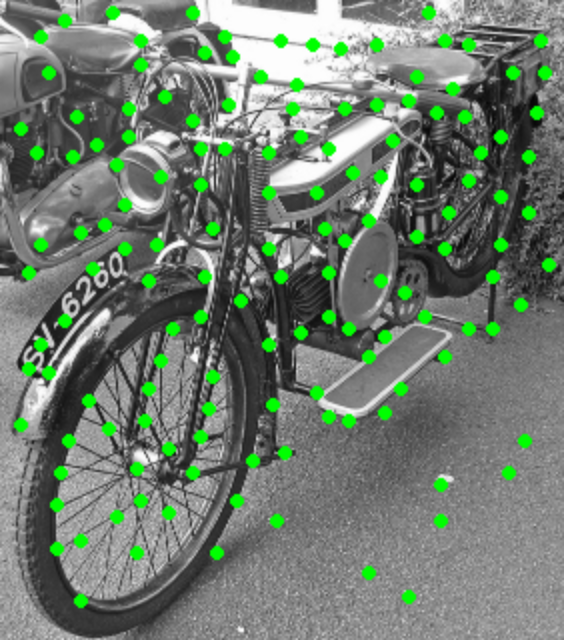} 

  \end{tabular}
  \caption{Comparison of keypoints.
  Left - superpoint points, right - goodpoint points. Each image has 143 points. As it can be seen many of goodpoint's points doesn't correspond to corners due to unsupervised learning, though many points coincide with corners.}
  \label{fig:compare}
\end{figure}
\begin{figure}[!t]
  \centering
  \begin{tabular}{@{}l l@{}}
    \includegraphics[scale=.295]{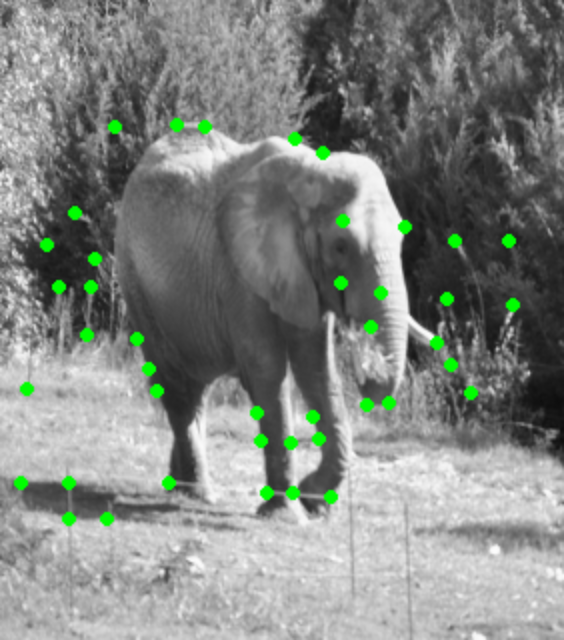} 
    & 
    \includegraphics[scale=.295]{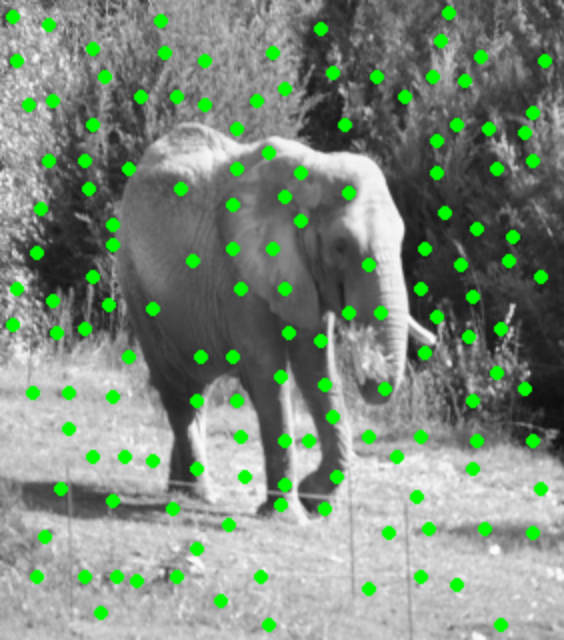} 

  \end{tabular}
  \caption{Comparison of keypoints.
  Left - superpoint points, right - goodpoint points. Left - superpoint points(57), right - goodpoint points(201). Same thresholds are used as for previous image.}
  \label{fig:compare1}
\end{figure}

Figures \ref{fig:compare} and \ref{fig:compare1} show side-by-side comparison of what networks tend to select as keypoints. The threshold is set so that in the first image networks detect the same number of keypoints. It can be seen that the  unsupervised model is less biased towards corner features, which may be an advantage or disadvantage depending on scene properties. More example images are available at the project website \footnote{https://github.com/singnet/image-matching}.

\subsection{AirSim village dataset}

AirSim village dataset was introduced in \citep{peterson2020distil}. It contains two sequences of images made with varying lighting but with the same camera positions. It contains ground truth camera pose and depth information, and as such may be used for the evaluation of SLAM or related methods, e.g. feature extraction and matching. Sequences were made by recording camera motion through a synthetic environment. The test was done on resolution 320x240 px. Matching was done with shift of 5 frames and radius of coverage set to 20 px. Precision and repeatability were calculated as average of matching in both ways: $I_i \to I_{i+5}$, and $I_{i+5} \to I_i$. The model was evaluated with and without roll of $5^\circ$. The original dataset wasn't generated with a roll in camera motion. 

The results are presented in table \ref{table:airsim}.  Threshold for correct match is 3 px. $\theta_{keypoint} = 0.028$ for GoodPoint, 0.015 for SuperPoint. $\theta_{desc}=0.8$ for both models. Overall, GoodPoint demonstrates good precision with lower than SuperPoint repeatability of keypoints.
\begin{figure*}[!t]
  \centering
  \begin{tabular}{@{}l l@{}}
    \includegraphics[scale=.35]{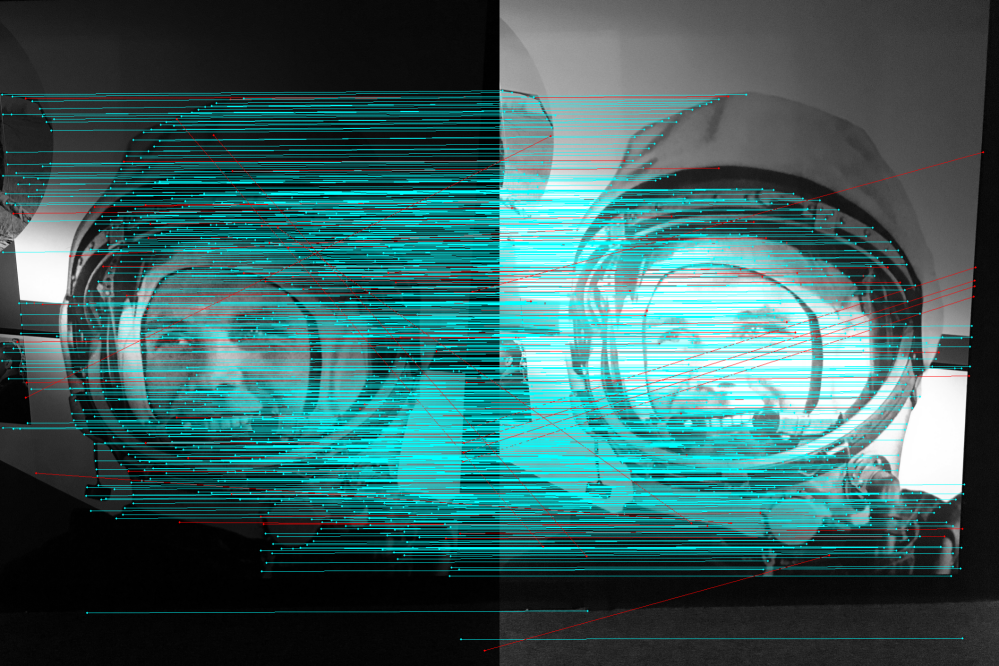} 
 &
    \includegraphics[scale=.35]{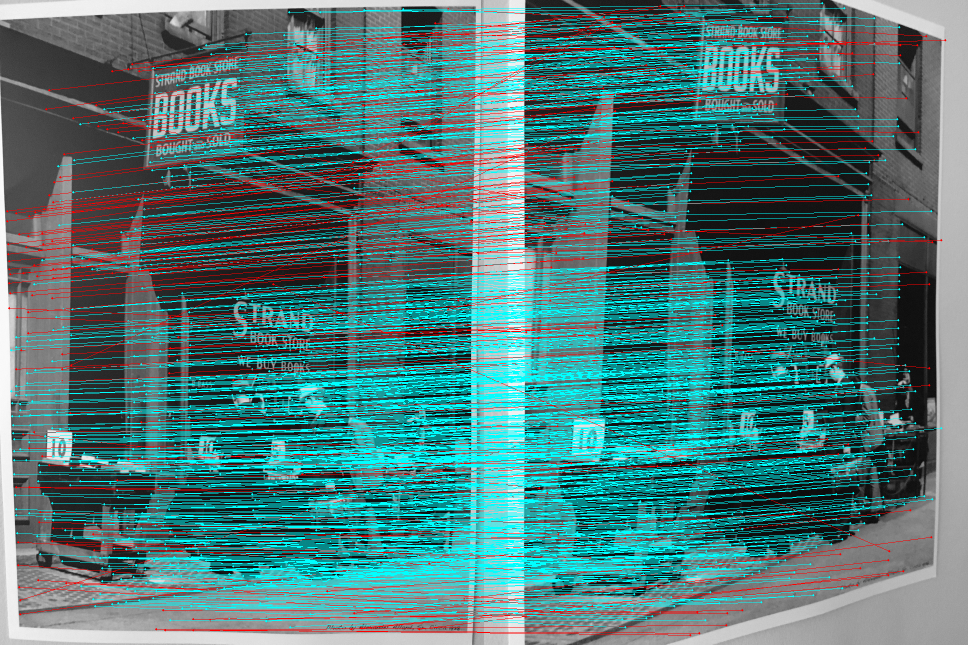} 

  \end{tabular}
  \vspace{\floatsep}

\begin{tabular}{@{}l l@{}}
    \includegraphics[scale=.35]{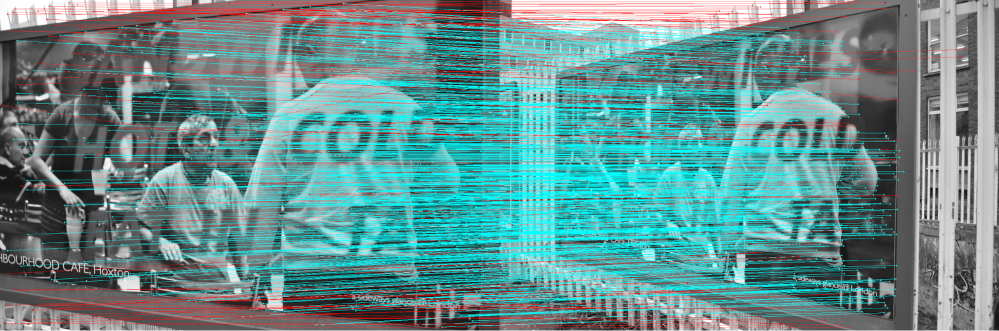} 
 &
    \includegraphics[scale=.35]{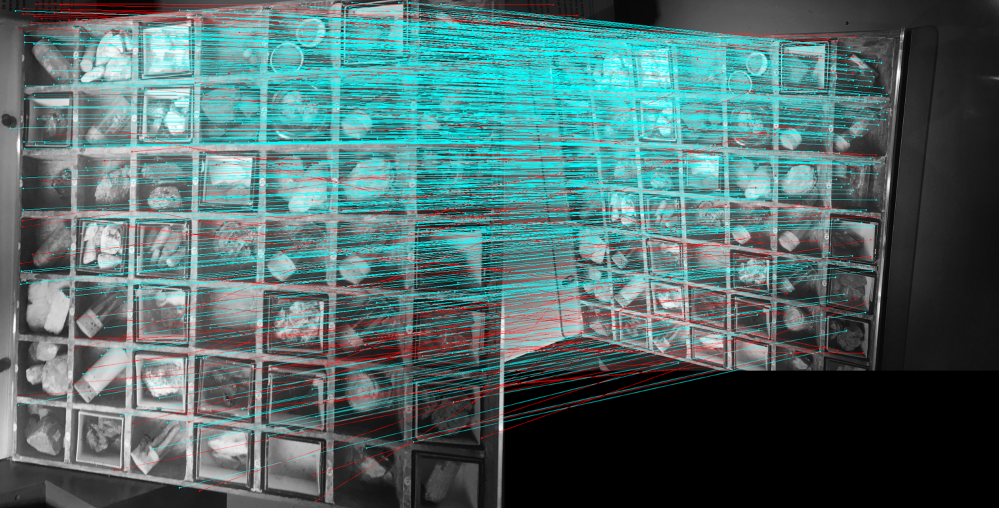} 

  \end{tabular}
  \caption{Points and matches from hpatches dataset}\label{fig:hpatches}
\end{figure*}

\subsection{HPatches}
For hpatches dataset the methodology of LF-net and SuperPoint papers have been used with thresholds for correct match set to 3 and 5 pixels. Coverage radius of 25px was used. The results are presented in table \ref{table:hpatches}.
The test demonstrates that the models have similar performance, with SuperPoint being more accurate in estimating keypoints positions, while GoodPoint tends to select more points, thus giving higher coverage, but lower replication ratio. 

\begin{table}[]
\caption{GoodPoint and SuperPoint on HPatches dataset}
\begin{tabular}{|l|l|l|l|l|}
\hline
\begin{tabular}[c]{@{}l@{}}
model\\ 
$\theta_{dist}$\\  
$\theta_{keypoint}$
\end{tabular} & \begin{tabular}[c]{@{}l@{}}GP\\ 3 px\\ 0.021\end{tabular} & \begin{tabular}[c]{@{}l@{}}SP\\ 3 px\\ 0.015\end{tabular} & \begin{tabular}[c]{@{}l@{}}GP\\ 5 px\\ 0.021\end{tabular} & \begin{tabular}[c]{@{}l@{}}SP\\ 5 px\\ 0.015\end{tabular} \\ \hline
Light Replication                                                & 0.48                                                      & 0.53                                                      & 0.63                                                      & 0.63                                                      \\ \hline
View Replication                                                 & 0.33                                                      & 0.45                                                      & 0.47                                                      & 0.55                                                      \\ \hline
Light Accuracy                                                   & 0.69                                                      & 0.70                                                      & 0.82                                                      & 0.8                                                       \\ \hline
View Accuracy                                                    & 0.53                                                      & 0.64                                                      & 0.67                                                      & 0.72                                                      \\ \hline
Light Coverage                                                   & 0.60                                                      & 0.47                                                      & 0.64                                                      & 0.50                                                      \\ \hline
View Coverage                                                    & 0.41                                                      & 0.42                                                      & 0.45                                                      & 0.45                                                      \\ \hline
Harmonic mean                                                    & 0.48                                                      & 0.52                                                      & 0.59                                                      & 0.59                                                      \\ \hline
\end{tabular}
\label{table:hpatches}
\end{table}
\subsection{Fundus Image Registration Dataset}

\begin{figure}[!t]
\centering
\includegraphics[scale=.75]{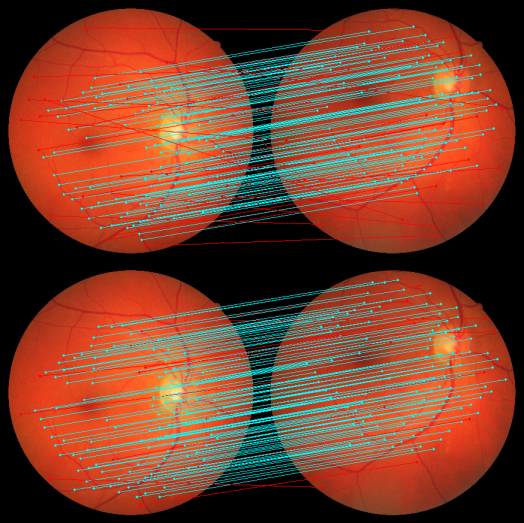} 
\caption{GoodPoint results on FIRE, top - before fine-tuning on FIRE dataset,
bottom - after.
}
\label{fig:fundus1}
\end{figure}

FIRE\citep{hernandez2017fire} dataset contains 134 pairs of retinal images with ground truth correspondences for a number of points, which allows for homography estimation. Also the dataset contains two masks, for global and local registration methods.  For this dataset, coverage radius is also set to 25 px. GoodPoint was tuned on images from FIRE, with the only change in the training pipeline being a different size of the crop window.
\begin{table}[]
\caption{Tests on FIRE dataset}
\begin{tabular}{|l|l|l|l|}
\hline
              & GoodPoint & \begin{tabular}[x]{@{}c@{}}GP tuned\\on FIRE\end{tabular} & SuperPoint \\ \hline
accuracy      & 0.78      & 0.79             & 0.84       \\ \hline
coverage      & 0.66      & 0.70             & 0.54       \\ \hline
replication   & 0.82      & 0.82             & 0.86       \\ \hline
harmonic mean & 0.75      & 0.77             & 0.71       \\ \hline
\end{tabular}
\label{table:fire}
\end{table}

Both original(trained on MS COCO) and fine-tuned versions were evaluated. The results are presented in table \ref{table:fire}. GoodPoint demonstrates better coverage than supervised SuperPoint, which shows that unsupervised learning of the keypoint detector introduced less bias into the model.

There is a trade-off between accuracy and coverage, and as shown in table \ref{table:fire075}, with the higher threshold for the keypoint detector it is possible to achieve the accuracy of GLAMPoint(0.91) on the FIRE dataset.  Coverage and replication ratio were not reported in the article \citep{truong2019glampoints}.  

\begin{table}[]
\caption{GoodPoint performance on FIRE with threshold = 0.075}
\begin{tabular}{|l|l|}
\hline
accuracy      & 0.91 \\
coverage      & 0.21 \\
replication   & 0.91 \\
harmonic mean & 0.44 
\\ \hline
\end{tabular}
\label{table:fire075}
\end{table}

\section{Conclusion and future work}
A novel method for joint training of keypoints detection and description has been introduced. The method is fully unsupervised and can be applied to train a model directly on a set of unlabelled images. The method was used to train convolutional model named GoodPoint. GoodPoint is based upon SuperPoint architecture. For the ease of comparison, only minor changes were introduced, such as removal of dustbin channel in keypoint detector, which was necessary for the proposed training method. As the result, GoodPoint has the same number of layers and parameters as SuperPoint. The trained model was evaluated on diverse datasets and demonstrated a good performance on natural and synthetic images, both rich(HPatches, AirSim village) and poor(FIRE) in corner features. GoodPoint tends to produce dense detections, which corresponds to higher coverage. The results open the way for the following improvements and/or research directions:
\begin{itemize}
\item
Replacement of maxpooling for keypoint extraction with theoretically sound sampling methods, such as $\epsilon$-greedy sampling.

\item
Augmenting local descriptors with global features, in the way it is done in SuperGlue during matching\citep{sarlin2019superglue}.
\end{itemize}

\bibliographystyle{model2-names.bst}
\bibliography{refs}

\end{document}